\documentclass{article} 
\usepackage{iclr2023_conference,times}

\usepackage{amsmath,amsfonts,bm}

\def\eqref#1{equation~\ref{#1}}

\def\1{\bm{1}}

\DeclareMathAlphabet{\mathsfit}{\encodingdefault}{\sfdefault}{m}{sl}
\SetMathAlphabet{\mathsfit}{bold}{\encodingdefault}{\sfdefault}{bx}{n}

\definecolor{mycitecolor}{rgb}{0, 0.4, 0.7}
\definecolor{myred1}{RGB}{196, 38, 38}
\definecolor{mygreen1}{RGB}{0, 153, 0}
\usepackage[breaklinks=true,bookmarks=true,citecolor=mycitecolor,colorlinks,pagebackref=true]{hyperref}
\usepackage{url}
\usepackage{graphicx}
\usepackage{booktabs}
\usepackage{multirow}
\usepackage{wrapfig,lipsum,booktabs}
\usepackage{multicol}
\usepackage{subcaption}
\usepackage{amsfonts}       
\usepackage{nicefrac}       
\usepackage{xcolor,colortbl}         
\usepackage{amsmath}        
\usepackage{textcomp, gensymb}
\usepackage[ruled,vlined]{algorithm2e}
\usepackage{enumitem, xspace}

\newcommand{\algname}{PPGeo\xspace}
\title{ 
Policy Pre-training for 
Autonomous Driving via Self-supervised Geometric Modeling}

\author{Penghao Wu$^{1,2}$, Li Chen$^1$, Hongyang Li$^{1,3}$\thanks{Hongyang Li is the correspondence author. This work was in part supported by NSFC (62206172, 62222607), Shanghai Municipal Science and Technology Major Project (2021SHZDZX0102), and Shanghai Committee of Science and Technology (21DZ1100100).}, Xiaosong Jia$^{3,1}$, Junchi Yan$^{3,1}$, Yu Qiao$^1$\\
$^1$Shanghai AI Laboratory\quad $^2$University of California at San Diego
\\$^3$MoE Key Lab of Artificial Intelligence, Shanghai Jiao Tong University\\
\texttt{\{wupenghao, lichen, lihongyang, qiaoyu\}@pjlab.org.cn}\\
\texttt{\{jiaxiaosong, yanjunchi\}@sjtu.edu.cn}\\ 
  \footnotesize \texttt{Code: \url{https://github.com/OpenDriveLab/PPGeo}}
}
\iclrfinalcopy 
\begin{document}

\maketitle

\begin{abstract}
Witnessing the impressive achievements of pre-training techniques on large-scale data in the field of computer vision and natural language processing, we wonder 
whether this idea could be adapted in a grab-and-go spirit, and
mitigate
the sample inefficiency problem for visuomotor driving.
%
Given the highly dynamic and variant nature of the input,
the visuomotor driving task inherently lacks 
view and translation invariance, and the visual input contains massive irrelevant information for decision making,
resulting in predominant pre-training approaches from general vision less suitable for the autonomous driving task.
%
To this end,
we propose \textbf{\algname} (Policy Pre-training via Geometric modeling), an intuitive and straightforward 
fully self-supervised framework curated for the policy pre-training in visuomotor driving.
We aim at learning policy representations as a powerful abstraction by modeling 3D geometric scenes on large-scale unlabeled and uncalibrated YouTube driving videos.
%
%
The proposed \algname is 
performed in two stages to support effective self-supervised training. In the first stage, the geometric modeling framework generates pose and depth
predictions simultaneously, with two consecutive frames as input. In the second stage, the visual encoder 
learns 
driving policy representation by predicting the future ego-motion and optimizing with the photometric error 
based on current visual observation only.
As such,
the pre-trained visual encoder is equipped with rich driving policy related representations and thereby 
competent for  multiple visuomotor driving tasks.
As a side product, 
the pre-trained geometric modeling networks could bring further improvement to the depth and odometry estimation tasks.
%
Extensive experiments 
covering a wide span of  challenging scenarios 
have demonstrated the superiority of 
our proposed approach, where improvements range from 2\% to even over 100\% with very limited data.
\end{abstract}

\section{Introduction}
\label{sec:intro}
\vspace{-4pt}
Policy learning refers to the learning process of an autonomous agent acquiring the  
decision-making policy to perform a certain task in a particular environment. Visuomotor policy learning~\citep{atari,levine2016end,hessel2018rainbow,laskin2020reinforcement,MaRLn} takes as input raw sensor observations and predicts the action, simultaneously cooperating and training the perception and control modules in an end-to-end fashion. For visuomotor policy models, learning tabula rasa is difficult, where it usually requires a prohibitively large corpus of labeled data or environment interactions to achieve satisfactory performance~\citep{impala,wijmans2019ddppo,yarats2020image}.

To mitigate the sample efficiency caveat in visuomotor policy learning, pre-training the visual perception network in advance is a promising solution. Recent studies~\citep{shah2021rrl,Parisi2022TheUE,xiao2022masked,radosavovic2022real,shah2022lmnav} have demonstrated that applying popular visual pre-training approaches, including ImageNet~\citep{deng2009imagenet} classification, contrastive learning~\citep{he2020moco,chen2020mocov2}, masked image modeling (MIM)~\citep{he2022mae,xie2022simmim,xue2022maskalign}, and language-vision pre-training~\citep{radford2021clip}, could guarantee superior representation for robotic policy learning tasks, \textit{e.g.},  dexterous manipulation, motor control skills and visual navigation. However, for one crucial and challenging visuomotor task in particular, namely end-to-end autonomous driving\footnote{We use \textit{end-to-end autonomous driving} and \textit{visuomotor autonomous driving} interchangeably in this paper.}, the aforementioned predominant pre-training methods may not be the optimal choice~\citep{yamada2022tarp, zhang2022aco}.

In this paper, we aim to investigate why ever-victorious pre-training approaches for general computer vision tasks and robotic control tasks are prone to \textit{fail} in case of  end-to-end autonomous driving. 
For conventional pre-training methods in general vision tasks, \textit{e.g.}, classification, segmentation and detection, they usually adopt a wide range of data augmentations
to achieve translation and view invariance~\citep{zhang2016colorization, wu2018npid}. 
For 
robotic control tasks, the input sequence is generally of small resolution; the environment setting is simple and concentrated on objects~\citep{Parisi2022TheUE, radosavovic2022real}.
We argue that the 
visuomotor driving investigated in this paper, is sensitive to geometric relationships and usually comprises complex scenarios.
%

\begin{figure}[tb!]
    \centering
    \includegraphics[width=0.98\textwidth]{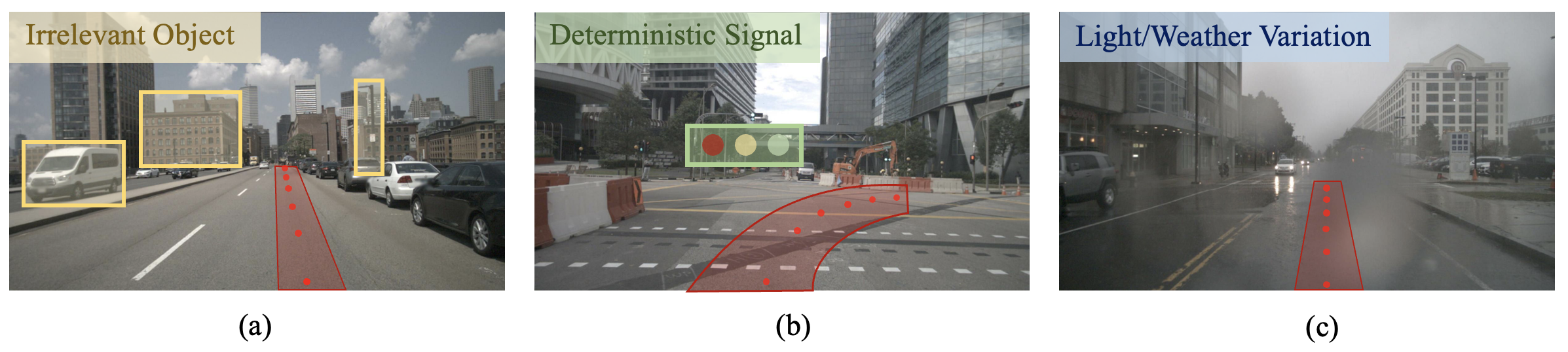}
    \vspace{-8pt}
    \caption{
    Uniqueness of visuomotor driving policy learning. The planned trajectory is shown as \textcolor{red}{red} points. \textbf{(a)} static obstacles
    and 
    background buildings 
    (objects in \textcolor{brown}{yellow} rectangles) are irrelevant to the driving decision; 
     \textbf{(b)} the traffic signal 
    in the visual input (marked with the \textcolor{mygreen1}{green} box) is extremely difficult to recognize and yet deterministic for control outputs;
    \textbf{(c)} the pre-trained visual encoder has to be robust to different light and weather conditions.  
    Photo credit from 
    \citep{caesar2020nuscenes}.}
    \label{fig:cases}
    \vspace{-10pt}
\end{figure}

As described in Fig.~\ref{fig:cases}(a), the input data 
often carry 
irrelevant information,
such as background buildings, far-away moving vehicles, nearby static obstacles, \textit{etc.}, 
which are deemed as noises for the decision making task.
%
To obtain a good driving policy, we argue that the desirable model should only concentrate on 
particular
parts/patterns of the visual input. 
That is,  taking heed of direct or deterministic relation to the decision making, \textit{e.g.}, traffic signals in Fig. \ref{fig:cases}(b).
However, concurrent pre-training approaches fail to fulfill such a requirement.
There comes a natural and necessary demand to formulate a pre-training scheme curated for end-to-end autonomous driving. 
We attempt to pre-train a visual encoder with a massive amount of driving data crawled freely from the web, 
such that given limited labeled data,
downstream applications could generalize well and quickly adapt to various driving environments as depicted in Fig. \ref{fig:cases}(c).

%
The pivotal question is \textit{how to introduce driving-decision awareness into the pre-training process to help the visual encoder concentrate on crucial visual cues for driving policy}.
%
One may resort to directly predicting ego-motion based on single frame sensor input, constraining the network on learning policy-related features. Previous literature tackles the supervision problem 
with pseudo labeling 
training on either an open dataset~\citep{zhang2022aco} or the target domain data~\citep{zhang2022selfd}.
However, pseudo labeling approaches suffer from noisy predictions from poorly calibrated models -
this is true especially when there exists distinct domain gap such as geographical locations and traffic complexities~\citep{rizve2020ups}.
%

To address the bottleneck aforementioned, 
we propose \textbf{\algname} (\textbf{P}olicy \textbf{P}re-training via \textbf{Geo}metric modeling), a fully self-supervised driving policy pre-training framework to learn from unlabeled and uncalibrated driving videos. 
It models the 3D geometric scene 
by jointly predicting ego-motion, depth, and camera intrinsics. 
Since directly learning ego-motion based on single frame input along with depth and intrinsics training from scratch is too difficult, it is necessary to separate the visual encoder pre-training from  depth and intrinsics learning in two stages.
In the first stage, the ego-motion is predicted based on consecutive frames as does in conventional depth estimation frameworks~\citep{godard17monodepth,godard19monodepth2}. In the second stage, the future ego-motion is estimated based on the single 
frame by a visual encoder, and could be optimized with the depth and camera intrinsics network well-learned  in the first stage. 
As such, the visual encoder is capable of inferring future ego-motion based on current 
input alone. 
%
The pre-trained visual encoder could be well adopted for downstream driving tasks since it captures
driving policy related information. 
As a side product, the depth and pose networks could be utilized as new initial weights for depth and odometry estimation tasks, bringing in an additional performance gain.
{To sum up, our \textbf{key contributions} are three-fold:}
\begin{itemize} [leftmargin=*,itemsep=0pt,topsep=0pt]
    \item We propose a 
    pre-training paradigm curated for various visuomotor driving tasks. To the best of our knowledge, this is the first attempt to achieve a fully self-supervised  
    framework \textit{without} any need of pseudo-labels\footnote{{Pseudo-labels here mean using another model trained on additional labeled data to create ``artificial'' labels for the unlabeled dataset.}}, leveraging the effect of pre-training by large-scale 
    data to the full extent. 
    \item We devise a visual encoder capable of predicting ego-motion 
    based on single visual input, being able to extract feature representations closely related to driving policy.
    Such a design of visual encoder is flexible to extend to various downstream applications.  
    %
    \item We demonstrate 
    the superiority of our approach on a set of end-to-end driving scenarios, covering different types and difficulty levels. The performance in terms of various metrics is improved from 2\% to even over 100\% in challenging cases with very limited data.
\end{itemize}


\section{Methodology}
\label{sec:method}
\vspace{-4pt}
\begin{figure}[tb!]
\centering
\includegraphics[width=0.88\textwidth]{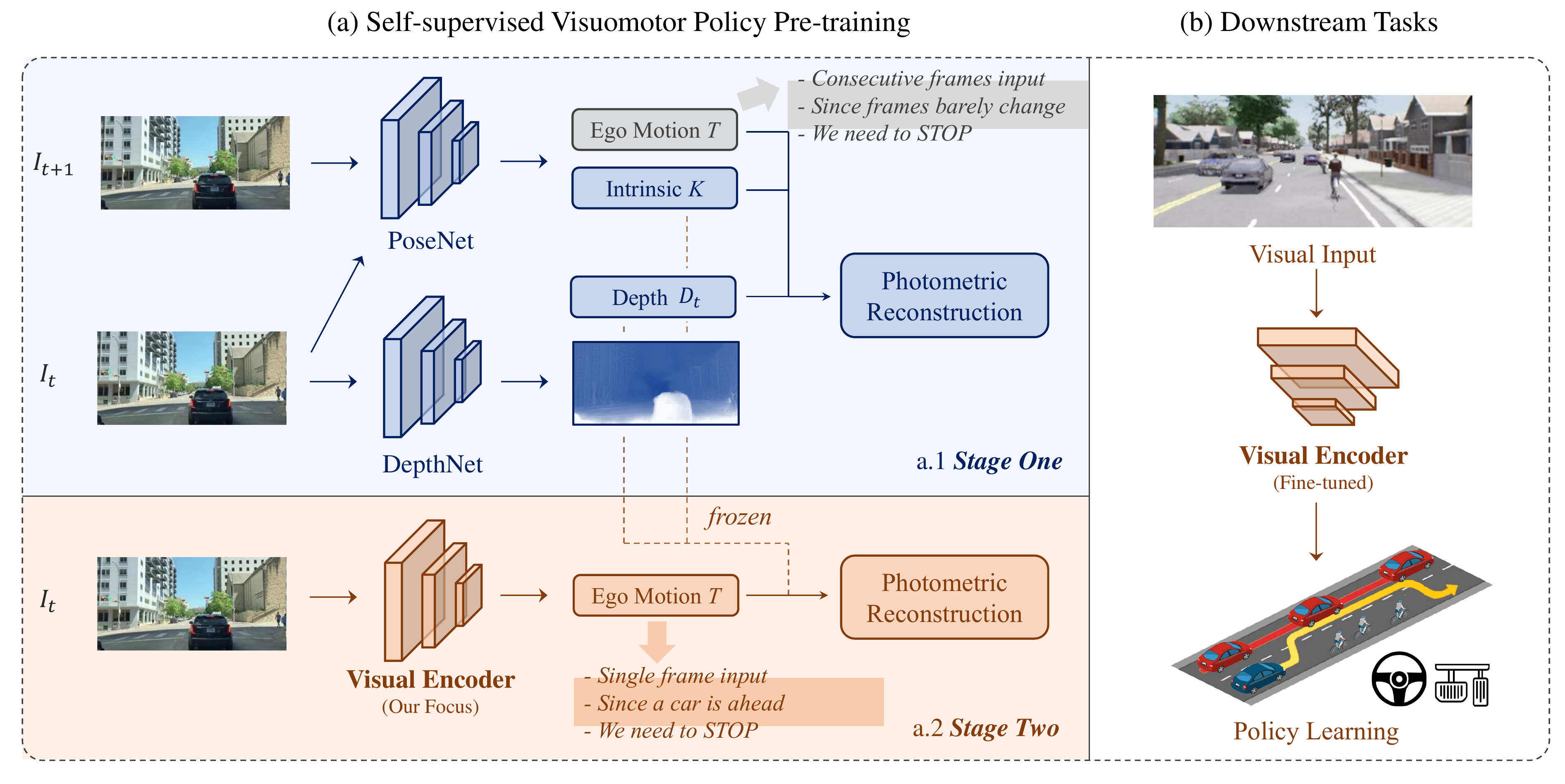}
\vspace{-8pt}
\caption{Overview of \textbf{\algname}. 
\textbf{(a)}
We focus on pre-training an effective visual encoder to encode driving policy related information by predicting ego-motion based on single frame input (a.2 Stage Two).
As achieving such a goal without labels is non-trivial, the visual encoder is obtained with the aid of a preceding procedure (a.1 Stage One) with temporal inputs and two sub-networks (pose and depth).
In this illustrative example, the ego-vehicle needs to take action of \texttt{STOP}. 
The ego-motion in (a.1) is inferred by judging two consecutive frames barely change; whilst the ego-motion in 
(a.2)
is predicted based on single visual input - focusing on driving policy related information.
%
%
As such, the visual encoder could be fine-tuned and applied to a wide span of downstream tasks in \textbf{(b)}. 
}
\label{fig:pipeline}
\vspace{-5pt}
\end{figure}

\subsection{Overview}
\vspace{-4pt}
The visuomotor policy learning for autonomous driving  targets generating a policy $\pi$, such that it makes driving decisions, \textit{e.g.}, 
control actions or planned trajectory, from visual observation $\mathbf{x}$. Our goal is to pre-train a visual encoder $\phi(\mathbf{x})$, which maps the raw image input to a compact representation containing important information for driving decision making. The representation is then utilized by the policy $\pi(\phi(\mathbf{x}))$ to perform driving tasks. As shown in Fig.~\ref{fig:pipeline}, our pre-training method pre-trains the visual encoder on unlabeled driving videos via two stages in a self-supervised manner.

\subsection{Two-stage Self-supervised Training}
\vspace{-4pt}
\textbf{Stage One: Self-supervised Geometric Modeling.} During the first stage, given a target image $I_t$ and source images $I_{t'}$ in a sequence, we jointly estimate the depth of the target image, the intrinsics of the camera, and the 6-DoF ego-motion between these two frames. Given the estimations, we are able to model the 3D geometry of the scene, and reconstruct the target image by projecting pixels in the source images.
Formally, 
the pixel-wise correspondence between $I_t$ and $I_{t'}$ is calculated as:
\begin{equation}
    \mathbf{p}_{t'} = \mathbf{KT}_{t\rightarrow t'}\mathbf{D}_t(\mathbf{p}_t)\mathbf{K}^{-1}\mathbf{p}_t,
\end{equation}
where $\mathbf{p}_t$ and $\mathbf{p}_{t'}$ are the homogeneous coordinates of the pixel in $I_t$ and $I_{t'}$ respectively, $\mathbf{K}$ is the predicted camera intrinsic matrix, and $\mathbf{D}_t(\mathbf{p}_t)$ represents the predicted depth value at pixel $p_i$ in $I_t$.
With this relationship, the target image $I_{t' \rightarrow t}$ could be reconstructed with pixels in $I_{t'}$, and be optimized by the photometric reconstruction error. Following~\citet{godard19monodepth2}, we choose two images adjacent to the current frame as the source images, \textit{i.e.}, $t' \in \{t-1, t+1\}$.

The DepthNet consists of a common encoder-decoder structure~\citep{godard19monodepth2} and estimates the depth map of the input image. Two images are stacked together and fed into the encoder of the PoseNet, whose bottleneck feature is then utilized to predict the camera intrinsics and the ego-motion via two separate MLP-based heads. For camera intrinsics estimation, optical center ($c_x, c_y$) and focal lengths $f_x, f_y$ are regressed similarly as in \citet{gordon2019depth, chanduri2021camlessmonodepth}.

\textbf{Stage Two: Visuomotor Policy Pre-training.} After the first stage of training, the DepthNet and PoseNet are well trained and fitted to the driving video data. Then, in the second stage, we replace the PoseNet for ego-motion estimation with the visual encoder {$\phi(\mathbf{x})$} prepared for downstream driving policy learning tasks. Now the visual encoder only takes a single frame image as input and predicts ego-motion between the current frame and subsequent frame. 

%
Specifically, the visual encoder estimates the ego-motion $T_{t\rightarrow t+1}$  based on $I_t$ alone and $T_{t\rightarrow t-1}$ based on $I_{t-1}$ followed by an inverse operation, respectively. 
The visual encoder is optimized by the photometric reconstruction error similar to the first stage, aside from a modification where the DepthNet and the intrinsics estimation 
are frozen and not backpropagated. This is empirically observed towards better performance. 
By doing so, the visual encoder is enforced to learn the actual driving policy, since the ego-motion between two consecutive frames is straightforwardly related to the driving decision or action taken at the current timestamp. 

One might argue that the PoseNet trained in the first stage could provide pseudo motion labels, with which the visual encoder could be directly supervised. However, the ego-motion predicted from the PoseNet is too sparse compared with the geometric projection approach. In our pipeline, every pixel provides supervision for the visual encoder so that inaccurate depth estimation in some pixels could be mitigated by the accurate ones, \textit{i.e.}, it constructs a ``global'' optimization.
In contrast, direct supervision from the PoseNet would be greatly affected by the undesirable prediction inaccuracy and noise results. 


Thus far, the {backbone} of visual encoder {$\phi(\mathbf{x})$} has gained knowledge about the driving policy from the diverse driving videos. It can then be applied to downstream visuomotor autonomous driving tasks as the initial weights. Besides, the DepthNet and PoseNet trained on this large corpus of uncalibrated video data could also be utilized in depth and odometry estimation tasks.

\subsection{Loss Function}
\vspace{-4pt}
Following~\citet{godard19monodepth2}, the loss function is comprised of the photometric loss and the smoothness loss. The photometric error is comprised of an $\ell_1$ term and an SSIM (structural similarity index measure) term~\citep{wang2004ssim}:
\begin{equation}
    \ell_{pe} = \frac{\alpha}{2}(1- \operatorname{SSIM}(I_t, I_{t'\rightarrow t})) + (1-\alpha)\ell_1(I_t, I_{t'\rightarrow t}),
\end{equation}
where we set $\alpha=0.85$ following the practice~\citep{godard17monodepth, godard19monodepth2}. The smooth loss is:
\begin{equation}
    \ell_s = |\partial_xd^*_t|e^{-|\partial_xI_t|} + |\partial_yd_t^*|e^{-|\partial_yI_t|},
\end{equation}
where $d^*_t$ is the mean-normalized inverse depth map. We also adopt the minimum reprojection loss and auto-masking scheme~\citep{godard19monodepth2} to improve self-supervised depth estimation.

\section{Experiments}
\label{sec:exp}
\vspace{-4pt}
All pre-training experiments are conducted on the hours-long unlabeled YouTube driving videos~\citep{zhang2022aco}. It covers different driving conditions \textit{e.g.}, geographical locations and weather.
We sample 0.8 million frames in total at 1 Hz for training.
For the first stage in \algname pipeline, we train the model for 30 epochs by Adam~\citep{Adam} optimizer with a learning rate of $10^{-4}$ which drops to $10^{-5}$ after 25 epochs.
For the second stage, the encoder is trained for 20 epochs using the AdamW~\citep{adamW} optimizer. A cyclic learning rate scheduler is applied with the learning rate ranging from $10^{-6}$ to $10^{-4}$.
The batch size for both stages is 128. We use data augmentations including ColorJitter, RamdomGrayScale, and GaussianBlur.

\subsection{Description on Compared Baselines}
\label{sec:exp-baseline}
\vspace{-4pt}
We use ResNet-34~\citep{he2016resnet} as the encoder and load different pre-trained weights for the initialization of downstream tasks. We compare \algname with pre-training methods including:

\textbf{Random.} We use the default Kaiming initialization~\citep{he2015delving} for convolution layers and constant initialization for batchnorms.  

\textbf{ImageNet.} We use the model weight provided by Torchvision~\citep{marcel2010torchvision}, which is pre-trained with the classification task on ImageNet~\citep{deng2009imagenet}. 

\textbf{MIM.} The model is pre-trained with the masked image modeling method on the YouTube driving video, which tries to reconstruct images with random masked-out patches. SimMIM~\citep{xie2022simmim} is adopted as it is suitable for convolutional networks. 

\textbf{MoCo.} We pre-train the model using MoCo-v2~\citep{chen2020mocov2} on the YouTube driving videos.  We exclude RandomResizedCrop and RandomHorizontalFlip augmentations as they are not suitable for the driving task.

\textbf{ACO.} Following~\citet{zhang2022aco}, it is pre-trained using action-conditioned contrastive learning on the YouTube driving videos. ACO trains an inverse dynamic model to generate pseudo steer labels for driving videos, based on which steer-based discrimination is added on top of MoCo-v2.

\textbf{SelfD.} SelfD~\citep{zhang2022selfd} is not a pre-training method strictly since it needs to train the whole policy model on the driving video for each task, while other pre-training methods aforementioned provide a general pre-training visual model for all tasks. We still include it for comparison due to its close relationship to our target. Specifically, we follow~\citet{zhang2022selfd} to train the model for each task with the following pipeline: training on the task data $\rightarrow$ training on the YouTube data with pseudo-label $\rightarrow$ fine-tuning on the task data.

\subsection{Description on Downstream Autonomous Driving Tasks}
\vspace{-4pt}
We carry out experiments under (1) three imitation learning based closed-loop driving tasks in CARLA~\citep{Dosovitskiy17carla}, (2) one reinforcement learning based driving task in CARLA, and (3) an open-loop planning task on real-world autonomous driving dataset nuScenes~\citep{caesar2020nuscenes}, to fully validate the effectiveness of \algname. We 
briefly describe each task below.

\textbf{Navigation.} It corresponds to the goal-conditioned navigation task in the CoRL2017 benchmark~\citep{Dosovitskiy17carla}. The agent is trained in Town01 and tested in Town02 with unseen weather, and there are no other traffic participants. We use different sizes of training data (from 4K to 40K) following~\citet{zhang2022aco} to evaluate the generalization ability of pre-trained visual encoders when labeled data is limited and conduct the closed-loop evaluation. The evaluation metric is success rate, denoting the portion of 50 pre-defined routes finished without any collision. And traffic lights are ignored here. CILRS~\citep{codevilla2019cilrs}, a classic image based end-to-end autonomous driving model, is adopted for training and evaluation.

\textbf{Navigation Dynamic.} 
This is the navigation dynamic task in the CoRL2017 benchmark~\citep{Dosovitskiy17carla}. The setting differentiates from Navigation that there are other dynamic objects such as randomly generated vehicles, which substantially increases the difficulty of driving safety.

\textbf{Leaderboard Town05-long.} This challenging and realistic benchmark corresponds to the LeaderBoard benchmark~\citep{carlaleaderboard}. We collect 40K training data in Town01, 03, 04, 06 and evaluate on 10 routes in the unseen Town05~\citep{prakash2021transfuser}. Due to the challenging scenarios in this task, we evaluate different pre-training approaches with the state-of-the-art image-based autonomous driving model TCP~\citep{wu2022tcp}. 
The metrics of this task are Driving Score, Route Completion, and Infraction Score (all the higher the better), where the main metric Driving Score is the product of Route Completion and Infraction Score.


\textbf{Reinforcement Learning.}
Proximal Policy Optimization (PPO)~\citep{schulman2017ppo} is used to train the CILRS~\citep{codevilla2019cilrs} model initialized with different pre-trained weights in CARLA Town01 environment. The reward shaping details follow Roach~\citep{zhang2021roach}.  We also conduct experiments to freeze the pre-trained visual encoder during training to further study the effectiveness of the pre-trained feature representations.

\textbf{nuScenes Planning.} 
This task involves trajectory planning in real-world dataset nuScenes~\citep{caesar2020nuscenes}. Given the current visual input, the model plans a 3-second trajectory (0.5 Hz), and the planned trajectory is compared with the ground truth log. We also calculate the collision rate, where a collision is defined as overlaps with future vehicles and pedestrians based on planned waypoints.
The planning model used here is comprised of a visual encoder and a GRU-based planner to predict each waypoint auto-regressively. We use the official train-val split for training and evaluation.

\subsection{Numeric Comparison on Downstream Tasks}
\begin{figure}[t!]
    \centering
    \includegraphics[width=\textwidth]{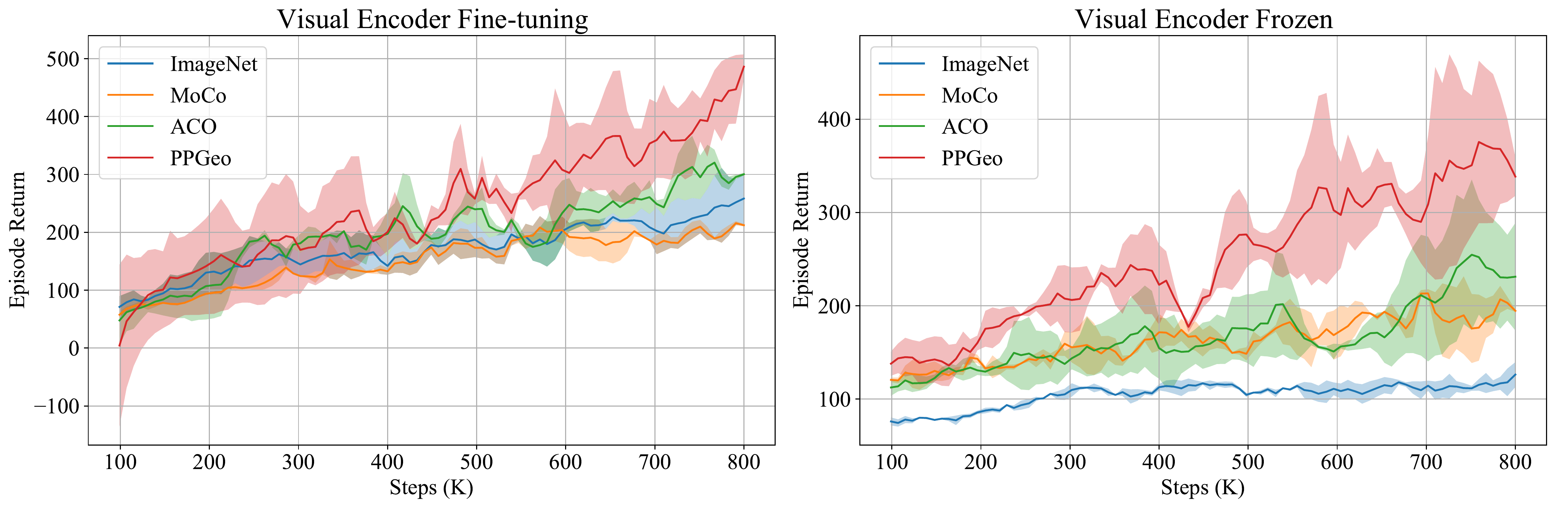}
    \vspace{-18pt}
    \caption{Learning curves of the RL agents using \algname and three other best pre-training baselines. Left: the pre-trained visual encoder is jointly fine-tuned during RL training; Right: the visual encoder is frozen during RL training. The episode return is the mean with standard deviation in shade across three runs with different random seeds.}
    \label{fig:rl}
    \vspace{-5pt}
\end{figure}

\vspace{-4pt}
For imitation learning based closed-loop driving tasks, the evaluation results are shown in Table~\ref{table:navigation}-\ref{table:town05long}. We present the plot between episode return and environment steps of each method in Fig.~\ref{fig:rl} for the reinforcement learning experiments. The open-loop nuScenes planning results are provided in Table~\ref{table:nuscenes planning}. We could observe that \algname outperforms other baselines by a large margin in \textit{all} tasks. 

Note that the model is tested under a different number of fine-tuning samples from 10\% (4K) to full 40K in the Navigation and Navigation Dynamic tasks. In the case of the particularly small size of training samples, \algname still demonstrates competitive performance and has a larger improvement gap of over 100\%. This validates the generalization ability of the pre-trained visual encoder, which is important when adapting to a new environment with very limited labeled data. In the more challenging and real-world style Leaderboard Town05-long task in Table~\ref{table:town05long}, the model pre-trained with our method achieves the highest driving score and infraction score. \algname well handles cases where the agent needs to stop, leading to much fewer vehicle collisions and red light infractions.


Since ACO considers steering angles only during pre-training, its performance degrades on more challenging scenarios where brake and throttles are also important. SelfD performs slightly better than ACO in complex cases while it significantly degenerates when the task data is limited, as affected by the unsatisfying pseudo labeling model.
ImageNet pre-training also shows competitive performance, which might credit to its ability of finding salient objects in the scene 
when the input contains little irrelevant information (see examples in Sec.~\ref{sec:visualization}).



\begin{table}[tb!]
\centering
\caption{The {Successful Rate}  of the closed-loop Navigation task  (mean by 3 random trials).}
\label{table:navigation}
\vspace{-8pt}
\scalebox{0.8}{
\begin{tabular}{@{}lcccc@{}}
\toprule
\multirow{2}{*}{Pre-train Method} & \multicolumn{4}{c}{Navigation - \# of training samples} \\ \cmidrule(l){2-5} 
 & 10\% (4K) & 20\% (8K) & 40\% (16K) & 100\% (40K) \\ \midrule
Random & 0.0 $\pm$ 0.0 & 9.6 $\pm$ 5.2 & 15.3 $\pm$ 4.5 & 73.3 $\pm$ 2.3 \\
ImageNet & 24.7$\pm$ 2.0 & 42.0 $\pm$ 2.0 & 69.3 $\pm$ 6.4 & 87.3 $\pm$ 4.6 \\
MIM & 4.7 $\pm$ 1.2 & 8.0 $\pm$ 0.0 & 31.3 $\pm$ 2.3 & 57.3 $\pm$ 3.1  \\
MoCo & 7.7 $\pm$ 2.1 & 39.3 $\pm$ 9.2  & 48.7 $\pm$ 4.2 & 69.3 $\pm$ 1.2 \\
ACO  & 24.0 $\pm$ 2.0 & 44.0 $\pm$ 1.2 & 71.3 $\pm$ 1.2 & 92.0 $\pm$ 3.5 \\
SelfD  &  12.0$ \pm$ 0.0  &   32.0 $\pm$ 0.0   &   50.7 $\pm$ 2.3   &   62.7 $\pm$ 1.2 \\
\midrule
\textbf{\algname} (ours) & \textbf{42.0 $\pm$ 2.0} & \textbf{73.3 $\pm$ 6.1} & \textbf{91.3 $\pm$ 1.2} & \textbf{96.7 $\pm$ 1.2} \\ \bottomrule
\end{tabular}}
\vspace{-5pt}
\end{table}

\begin{table}[t!]
\centering
\caption{The {Successful Rate}  of the closed-loop Navigation Dynamic (mean by 3 random trials).}
\vspace{-2pt}
\label{table:nav dynamic}
\begin{tabular}{@{}lcccc@{}}
\toprule
\multirow{2}{*}{Pre-train Method} & \multicolumn{4}{c}{Navigation Dynamic - \# of training samples} \\ \cmidrule(l){2-5} 
 & 10\% (4K) & 20\% (8K) & 40\% (16K) & 100\% (40K) \\ \midrule
Random & 0.0 $\pm$ 0.0 & 2.0 $\pm$ 0.0 & 10.0 $\pm$ 0.0 & 32.0 $\pm$ 8.0 \\
ImageNet & 10.7$\pm$ 1.2 & 28.7 $\pm$ 5.0 & 64.7 $\pm$ 2.3 & 72.7 $\pm$ 1.2 \\
MIM & 7.3 $\pm$ 1.2 & 10.3 $\pm$ 2.5 & 14.7 $\pm$ 3.1 & 58.7 $\pm$ 1.2  \\
MoCo & 4.7 $\pm$ 1.2 & 12.0 $\pm$ 4.0  & 28.0 $\pm$ 5.3 & 66.7 $\pm$ 2.3 \\
ACO  & 8.0 $\pm$ 1.2 & 12.0 $\pm$ 0.0 & 22.0 $\pm$ 2.0 & 47.3 $\pm$ 5.0 \\
SelfD  &  8.0 $\pm$ 0.0 &  29.3 $\pm$ 1.2   &  38.0 $\pm$ 1.6   &   59.3 $\pm$ 6.4 \\
\midrule
\textbf{\algname} (ours) & \textbf{23.3 $\pm$ 1.2} & \textbf{34.0 $\pm$ 5.3} & \textbf{71.3 $\pm$ 1.2} & \textbf{84.0 $\pm$ 5.3} \\ \bottomrule
\end{tabular}
\vspace{-2pt}
\end{table}

\begin{table}[t!]
\centering
\caption{Closed-loop Leaderboard Town05-long task results. Besides three main metrics, infraction details are also reported (all the lower the better). Evaluation repeats 3 times with the mean reported. }
\label{table:town05long}
\vspace{-8pt}
\scalebox{0.7}{
\begin{tabular}{@{}l>{\columncolor[gray]{0.9}}ccccccccc@{}}
\toprule
\begin{tabular}[c]{@{}c@{}}Pre-train \\ Method\end{tabular} & \textbf{\begin{tabular}[c]{@{}c@{}}Driving \\ Score\end{tabular}} & \begin{tabular}[c]{@{}c@{}}Infraction \\ Score\end{tabular} & \begin{tabular}[c]{@{}c@{}}Route \\ Completion\end{tabular} & \begin{tabular}[c]{@{}c@{}}Collisions \\ pedestrian\end{tabular} & \begin{tabular}[c]{@{}c@{}}Collisions \\ vehicle\end{tabular} & \begin{tabular}[c]{@{}c@{}}Collisions \\ layout\end{tabular} & \begin{tabular}[c]{@{}c@{}}Off-road \\ violations\end{tabular} & \begin{tabular}[c]{@{}c@{}}Agent \\ blocked\end{tabular} & \begin{tabular}[c]{@{}c@{}}Red light \\ violations\end{tabular} \\ \midrule
Random & 33.50$\pm$1.67 & 0.65$\pm$0.02 & 60.49$\pm$2.93 & 0.09$\pm$0.07 & 1.16$\pm$0.40 & 0.00$\pm$0.00 & 0.44$\pm$0.13 & 0.97$\pm$0.09 & 0.53$\pm$0.12 \\
ImageNet & 41.29$\pm$3.20 & 0.77$\pm$0.03 & 57.52$\pm$4.87& 0.00$\pm$0.00 & 0.71$\pm$0.20 & 0.11$\pm$0.15 & 0.15$\pm$0.01 & 1.01$\pm$0.16 & 0.29$\pm$0.10  \\
MIM & 36.39$\pm$0.21 & 0.72$\pm$0.04 & 61.75$\pm$2.26 & 0.14$\pm$0.11 & 0.91$\pm$0.12 & 0.04$\pm$0.07 & 0.18$\pm$0.17 & 0.87$\pm$0.03 & 0.14$\pm$0.11 \\
MoCo & 32.10$\pm$2.04 & 0.65$\pm$0.02 & 64.09$\pm$4.01& 0.13$\pm$0.11  & 0.79$\pm$0.16 & 0.00$\pm$0.00 & 0.49$\pm$0.07& 0.81$\pm$0.15& 0.45$\pm$0.13 \\
ACO & 33.05$\pm$3.05 & 0.67$\pm$0.06 & 59.52$\pm$3.21 & 0.00$\pm$0.00 & 0.69$\pm$0.28 & 0.05$\pm$0.07 & 0.54$\pm$0.05 & 0.94$\pm$0.08 &0.73$\pm$0.10  \\
SelfD & 38.76$\pm$3.02 & 0.65$\pm$0.03 & \textbf{68.72$\pm$7.36} & 0.17$\pm$0.07 & 0.84$\pm$0.18  & 0.00$\pm$0.00 & 0.32$\pm$0.03 & 0.75$\pm$0.15 & 0.12$\pm$0.08  \\
 \midrule
\textbf{\algname} & \textbf{47.44$\pm$5.63} & \textbf{0.79$\pm$0.08}  & 65.05$\pm$5.11 & 0.04$\pm$0.05 & 0.54$\pm$0.29 & 0.00$\pm$0.00 & 0.16$\pm$0.11 & 0.76$\pm$0.10 &0.04$\pm$0.05 \\ \bottomrule
\end{tabular}}
\vspace{-5pt}
\end{table}

\begin{table}[t!]
\centering
\caption{Open-loop nuScenes planning results. We evaluate the $\ell_2$ distance between model predictions and the ground truth trajectory and collision rate in horizons from 1 second to 3 seconds.}
\vspace{-8pt}
\label{table:nuscenes planning}
\scalebox{0.85}{
\begin{tabular}{@{}lcccccc@{}}
\toprule
\multirow{2}{*}{Pre-train Method} & \multicolumn{3}{c}{{L2 (m) $\downarrow$}} & \multicolumn{3}{c}{Collision Rate (\%) $\downarrow$} \\ \cmidrule(lr){2-4} \cmidrule(l){5-7} 
& 1s& 2s & 3s & 1s& 2s & 3s  \\ \midrule
Random & 1.621 & 2.722  & 3.851 & 0.550 & 1.779 & 3.375  \\
ImagNet  & 1.331& 2.202 & 3.086 & 0.315 & 0.550 & 1.366\\ 
MIM  & 1.412 & 2.357 & 3.331   & 0.297 & 0.622 & 1.507\\
MoCo & 1.528  & 2.545 & 3.585  & 0.560 & 1.235 & 2.390\\
ACO & 1.496 & 2.496 & 3.519 & 0.446 & 1.178 & 2.223 \\
SelfD & 1.419 & 2.359 & 3.316 & 0.353 & 0.923 & 2.044 \\ \midrule
\textbf{\algname} (ours)  & \textbf{1.302}  & \textbf{2.154}  & \textbf{3.018} & \textbf{0.270}  & \textbf{0.425}  & \textbf{0.941} \\ \bottomrule
\end{tabular}}
\vspace{-5pt}
\end{table}

\begin{table}[tb!]
\centering
\caption{Improvement from our pre-training method on depth and odometry estimation tasks.}\vspace{-8pt}
\label{table:depth odometry}
\scalebox{0.82}{
\begin{tabular}{@{}lccccccccc@{}}
\toprule
 \multirow{2}{*}{\begin{tabular}[c]{@{}c@{}}Pre-train \\ Method\end{tabular}} & \multicolumn{7}{c}{Depth Estimation}& \multicolumn{2}{c}{Odometry Estimation} \\ \cmidrule(r){2-8}  \cmidrule(l){9-10}
 & abs\_rel $\downarrow$ & sq\_rel $\downarrow$ & rmse $\downarrow$  & rmse\_log $\downarrow$ & a1 $\uparrow$   & a2 $\uparrow$  & a3 $\uparrow$ & Sequence 09  $\downarrow$ & Sequence 10 $\downarrow$ \\ \cmidrule(r){1-8} \cmidrule(l){9-10}
ImageNet  & 0.118    & 0.902   & 4.873 & 0.196 & 0.871  & 0.958 & 0.981 & 0.017$\pm$0.010 & 0.015$\pm$0.010     \\
\textbf{\algname}  & \textbf{0.114}    & \textbf{0.805}   & \textbf{4.599} & \textbf{0.186}     & \textbf{0.874} & \textbf{0.962} & \textbf{0.984} & \textbf{0.016$\pm$0.009} & \textbf{0.013$\pm$0.009}  \\ \bottomrule
\end{tabular}}
\vspace{-5pt}
\end{table}

%

\subsection{Depth and Odometry Estimation}
\vspace{-4pt}
In this part, we explore whether the large-scale training on uncalibrated data could benefit the depth and odometry estimation models as well and validate the effectiveness of first-stage training.
Specifically, we employ the DepthNet and PoseNet trained after the first stage as initial weights for Monodepthv2~\citep{godard19monodepth2}, and conduct experiments on KITTI~\citep{Geiger2012kitti}.
Results in Table~\ref{table:depth odometry} indicate that pre-training on large-scale driving videos could bring performance improvement to both depth and odometry estimation tasks, which is an additional harvest of our pre-training framework. We refer readers to \citet{godard19monodepth2} for details about the metrics of these tasks.

\subsection{Visualization Results}
\label{sec:visualization}
\vspace{-4pt}

\begin{figure}[tb!]
    \centering
    \includegraphics[width=0.85\textwidth]{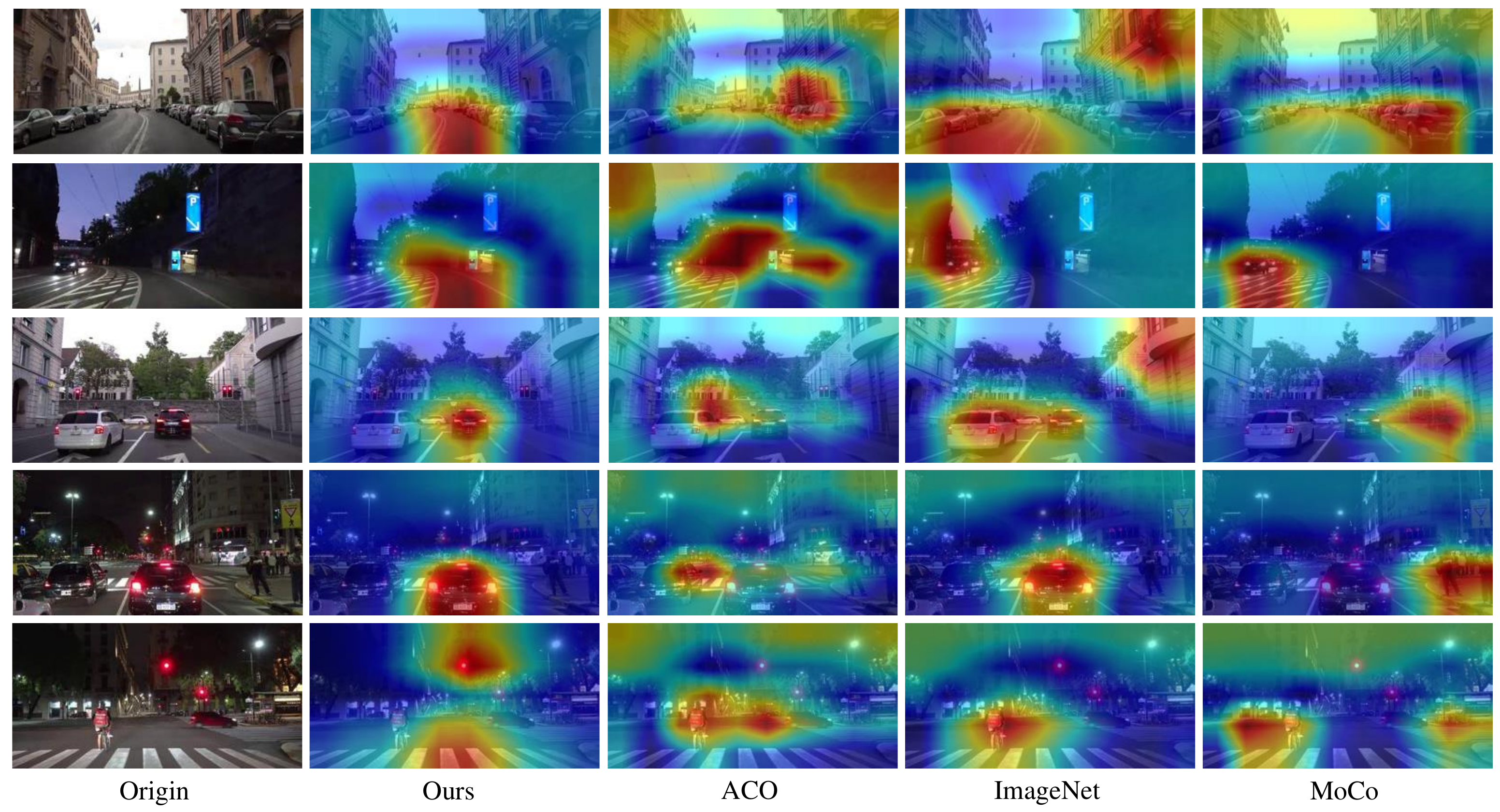}
    \vspace{-5pt}
    \caption{Eigen-Cam~\citep{eigencam} activation maps of the learned representation from different pre-training methods on the driving video data. }
    \label{fig:heatmap}
    \vspace{-5pt}
\end{figure}

Here we provide heatmaps of the feature representations learned by different pre-training methods using Eigen-Cam~\citep{eigencam} to show the attended regions in Fig.~\ref{fig:heatmap}. In many cases (Row 1\&2), our model mainly concentrates on the lane in front of the ego vehicle, which is highly related to driving. 
And our model  PPGeo well captures the specific cues causing the brake action including front vehicles (Row 3\&4) and traffic lights (Row 5). We also observe that the model pre-trained with ImageNet classification tends to capture salient objects in the image. This is helpful when the salient objects are straightforwardly related to the driving decision (Row 4); but it may focus on wrong objects when the input contains other irrelevant information (Row 2\&3).


\subsection{Ablative Study}
\vspace{-4pt}
\begin{table}[t!]
\centering
\caption{Ablative study on key designs of \algname on the Navigation task.}
\vspace{-8pt}
\label{table:ablation}
\scalebox{0.8}{
\begin{tabular}{@{}clcccc@{}}
\toprule
\multirow{2}{*}{\#} &
\multirow{2}{*}{Experiment} & \multicolumn{4}{c}{Navigation - \# of training samples} \\ \cmidrule(l){3-6} 
\multicolumn{1}{c}{}       & \multicolumn{1}{c}{}   & 10\% (4K)    & 20\% (8K)    & 40\% (16K)  & 100\% (40K) \\ \midrule
1& Single stage                                   & 24.2 $\pm$ 2.0  & 53.3 $\pm$ 1.2  & 79.3 $\pm$ 4.2 & 92.7 $\pm$ 2.3 \\
2& No frozen in 2nd stage                         & 32.7 $\pm$ 1.2  & 58.0 $\pm$ 2.0  & 86.0 $\pm$ 2.1 & 92.0 $\pm$ 2.0 \\
3& PoseNet direct supervision                     & 18.0 $\pm$ 2.0             &  52.0 $\pm$ 2.0            &     76.7 $\pm$ 1.2        &   90.0 $\pm$ 0.0          \\ \midrule
4&\textbf{\algname}                                          & 42.0 $\pm$ 2.0  & 73.3 $\pm$ 6.1  & 91.3 $\pm$ 1.2 & 96.7 $\pm$ 1.2 \\ \bottomrule
\end{tabular}}
\vspace{-5pt}
\end{table}

We conduct ablative study as to different designs of \algname on the Navigation task in Table~\ref{table:ablation}. Training the visual encoder and DepthNet in a single stage simultaneously (Row 1) leads to an inferior performance, indicating that it is quite challenging for the visual encoder to learn the correct ego-motion if depth estimation is also trained from scratch.
Moreover, jointly optimizing the DepthNet in the second stage (Row 2, not frozen) degrades the depth estimation quality and harms the performance.
In Row 3, we observe that utilizing the PoseNet obtained in the first stage to provide pseudo label supervision directly leads to inferior results, since an inaccurate pseudo label impairs the learning process to great extent.

\section{Related Work}
\label{sec:rel}
\vspace{-5pt}

\textbf{Pre-training for NLP and General Vision.}
Pre-training or representation learning has proved to be an essential key to the success of artificial intelligence. In the field of Natural Language Processing (NLP), with the powerful capability of Transformer~\citep{vaswani2017transformer}, pre-training on large-scale datasets with large models then fine-tuning on downstream tasks has become the dominant paradigm~\citep{kenton2019bert,brown2020gpt3}.
As for the field of Computer Vision, training specific downstream tasks with the supervised pre-trained weights of visual encoder via ImageNet classification task is widely adopted.
Recently, unsupervised and self-supervised learning methods such as contrastive learning~\citep{he2020moco,chen2020mocov2, simCLR} and masked image modeling~\citep{bao2021beit,he2022mae,xie2022simmim,peng2022beitv2,xue2022maskalign} have gained impressive improvement over ImageNet pre-training on various vision benchmarks.
Very recent vision-language co-training approaches~\citep{radford2021clip,wang2022beit3} demonstrate their extraordinary potential in the domain of multi-modal learning and applications.
Yet, these generic representation learning methods adopt various data augmentation techniques to achieve translation and view invariance, while visuomotor driving sets in a highly dynamic environment.
In this work, we show that the ever-victorious pre-training methods may not be the optimal choice, and introduce a curated paradigm for visuomotor driving policy learning.

\textbf{Pre-training for Visuomotor Applications.}
Learning a control policy directly from raw visual input is challenging since the model needs to reason about visual pixels and dynamic behaviors simultaneously. Moreover, training visuomotor models from scratch usually requires tons of labeled data or environment interactions.
To this end, recently, \citet{shah2021rrl} shows that feature representations from ResNet~\citep{he2016resnet} pre-trained on ImageNet classification is helpful for RL-based dexterous manipulation tasks.
\citet{Parisi2022TheUE} conducts extensive experiments on applying ``off-the-shelf'' pre-trained vision models in diverse control domains and validates their benefits to train control policies.
CLIP~\citep{radford2021clip} is also adopted in some embodied AI and robot navigation problems~\citep{shah2022lmnav}.
Besides borrowing pre-trained weights for visuomotor tasks, researchers in robotics now desire a paradigm learning policy representations from raw data directly.
\citet{xiao2022masked,radosavovic2022real,seo2022mwm, gupta2022maskvit} inherit the MIM spirit to realize visual pre-training for control tasks.
\citet{yang2021representation} investigates unsupervised representation learning objectives from D4RL environments~\citep{fu2020d4rl}, and \citet{yamada2022tarp} further adopts task-induced approaches to learn from prior tasks. However, compared with visuomotor driving, the visual inputs of such control tasks are less diverse which usually concentrate on objects and are much more compact.

To our best knowledge, ACO~\citep{zhang2022aco} is the only pre-training method customized for autonomous driving. By first training an inverse dynamic model on nuScenes~\citep{caesar2020nuscenes}, they get pseudo steer labels of the driving videos and then construct the steer-conditioned discrimination  for contrastive learning following MoCo. However, ACO ignores other crucial driving factors such as throttle and brakes, and its performance is largely limited by the inverse dynamic model.
SelfD~\citep{zhang2022selfd} is not strictly designed for pre-training while it also makes use of vast amounts of videos  to learn driving policies via semi-supervised learning. It acquires the pseudo labeling knowledge from the target domain. 
These two methods both depend on the accuracy of pseudo labeling. In contrast, we realize fully self-supervised learning through dense geometric reconstruction, evading the possible adverse effect.

\textbf{Policy Learning for Autonomous Driving.}
Visuomotor autonomous driving learns a driving policy directly from sensor inputs in an end-to-end manner~\citep{codevilla2018cil,codevilla2019cilrs,liang2018cirl,chen2020lbc,prakash2021transfuser,chen2021learning,wu2022tcp,shao2022interfuser,hu2022stp3,uniad}.
In essence, the inherent difficulty of the urban-style autonomous driving tasks makes such methods data-hungry. Interfuser~\citep{shao2022interfuser}, the current top-1 method on the CARLA Leaderboard~\citep{carlaleaderboard}, requires 3 million labeled data samples for imitation learning (behavior cloning specifically). RL-based model MaRLn~\citep{MaRLn} needs 20 million environment steps of interaction.
The sample efficiency problem greatly impedes the real-world application of such approaches.
In this work, we propose a self-supervised pre-training pipeline to learn driving policy related representations on unlabeled driving videos, and pave the way for these visuomotor autonomous driving models to further achieve satisfying performance.
%

\section{Conclusion}
\label{sec:conc}
\vspace{-5pt}
In this work, we have proposed a fully self-supervised visuomotor driving policy pre-training paradigm \algname by modeling the 3D geometry of large-scale unlabeled driving videos.
Taking a direct approach to infer the ego-motion and benefiting from the two-stage pre-training pipeline, we enable the visual encoder to learn driving policies based on single visual input. 
Our method outperforms the peer pre-training approaches by a large margin on a series of visuomotor driving tasks.
%


\bibliography{iclr2023_conference}
\bibliographystyle{iclr2023_conference}
\clearpage

\appendix

\section*{\Large {
Policy Pre-training for 
Autonomous Driving\\via Self-supervised Geometric Modeling\\
\textit{Supplementary Materials}}}



%


In this Supplementary document, we first provide detailed network structures in Sec.~\ref{sec:network details}. More description and visual illustrations of the downstream tasks are discussed in Sec.~\ref{sec:task details}. Last, we discuss limitations and common failure cases in Sec.~\ref{sec:limitation}.

\section{Network Details} \label{sec:network details}

For all experiments, the backbone of the visual encoder is ResNet-34~\citep{he2016resnet}, and the detailed structure of it is provided in Table~\ref{table:visual_encoder_detail}. For DepthNet and PoseNet, we follow the same model structure as~\citet{godard19monodepth2} with a two-layer MLP focal length head and a two-layer MLP optical center head added to the bottleneck of the PoseNet to predict the  intrinsic matrix. Please refer to~\citet{godard19monodepth2} for model details. 

For the Navigation, Navigation Dynamic, and Reinforcement Learning tasks, we use CILRS~\citep{codevilla2019cilrs} and the model details are provided in Table~\ref{table:cilrs_structure}. For the Leaderboard Town05-long task, TCP~\citep{wu2022tcp} is chosen as our agent, and we refer readers to~\citet{wu2022tcp} for model details. For the 
nuScenes Planning, the trajectory planning model structure is shown in Table~\ref{table:trajectory_structure}.

\begin{table}[hbt!]
\centering
\caption{Detailed structure of the visual encoder.}
\begin{tabular}{@{}lcccc@{}}
\toprule
Layer Type  & Channels & Stride & Kernel Size  & Activation Function  \\ \midrule
\multicolumn{5}{c}{\textbf{Image Encoder}}                         \\ \midrule
ResNet-34    &          &         &                     &    \\ \midrule
\multicolumn{5}{c}{\textbf{Measurement Encoder}}                   \\ \midrule
Conv          & 256               & 1 & 1                & ReLU  \\ 
Conv          & 256               & 3 & 1                & ReLU  \\
Conv          & 256               & 3 & 1                & ReLU  \\
Conv          & 6               & 1 & 1                & ReLU  \\ 
Average Pooling &          &         &                     & \\
\bottomrule
\end{tabular}
\label{table:visual_encoder_detail}
\end{table}

\begin{table}[hbt!]
\centering
\caption{Detailed structure of the CILRS model.}
\begin{tabular}{@{}lccc@{}}
\toprule
Layer Type  & Dims in & Dims out     & Activation Function   \\ \midrule
\multicolumn{4}{c}{\textbf{Image Encoder}}                         \\ \midrule
ResNet-34    &                   &    512                 &    \\ \midrule
\multicolumn{4}{c}{\textbf{Speed Encoder}}                   \\ \midrule
FC          & 1  & 256              & ReLU       \\
FC          & 256 & 512             & -         \\ \midrule
\multicolumn{4}{c}{\textbf{Speed Pred Head}}                            \\ \midrule
FC          & 512 & 256              & ReLU               \\
FC          & 256 & 256              & ReLU                \\ \midrule
FC          & 256 & 256              & ReLU                \\ \midrule
\multicolumn{4}{c}{\textbf{Control Pred Head}}                            \\ \midrule
FC          & 512 & 256            & ReLU              \\
FC          & 256 & 256              & ReLU               \\ 
FC          & 256 & 3              & Sigmoid               \\ 
 \bottomrule
\end{tabular}
\label{table:cilrs_structure}
\end{table}

\begin{table}[hbt!]
\centering
\caption{Detailed structure of the trajectory planning model.}
\begin{tabular}{@{}lccc@{}}
\toprule
\multicolumn{4}{c}{\textbf{Image Encoder}}                         \\ \midrule
ResNet-34    &                   &                     &    \\ \midrule
\multicolumn{4}{c}{\textbf{Bottleneck}}                   \\ \midrule
Layer Type  & Dims in & Dims out     & Activation Function   \\ \midrule
FC          & 512  & 256              & ReLU       \\
FC          & 256 & 256             & -         \\ \midrule
\multicolumn{4}{c}{\textbf{Decoder}}                            \\ \midrule
Layer Type  & Hidden dim & Input Dim & Output Dim   \\ \midrule
GRU & 256 & 2 &2 \\
 \bottomrule
\end{tabular}
\label{table:trajectory_structure}
\end{table}

\section{Downstream Tasks Details} \label{sec:task details}

\begin{figure}[htb]
    \centering
    \includegraphics[width=0.6\textwidth]{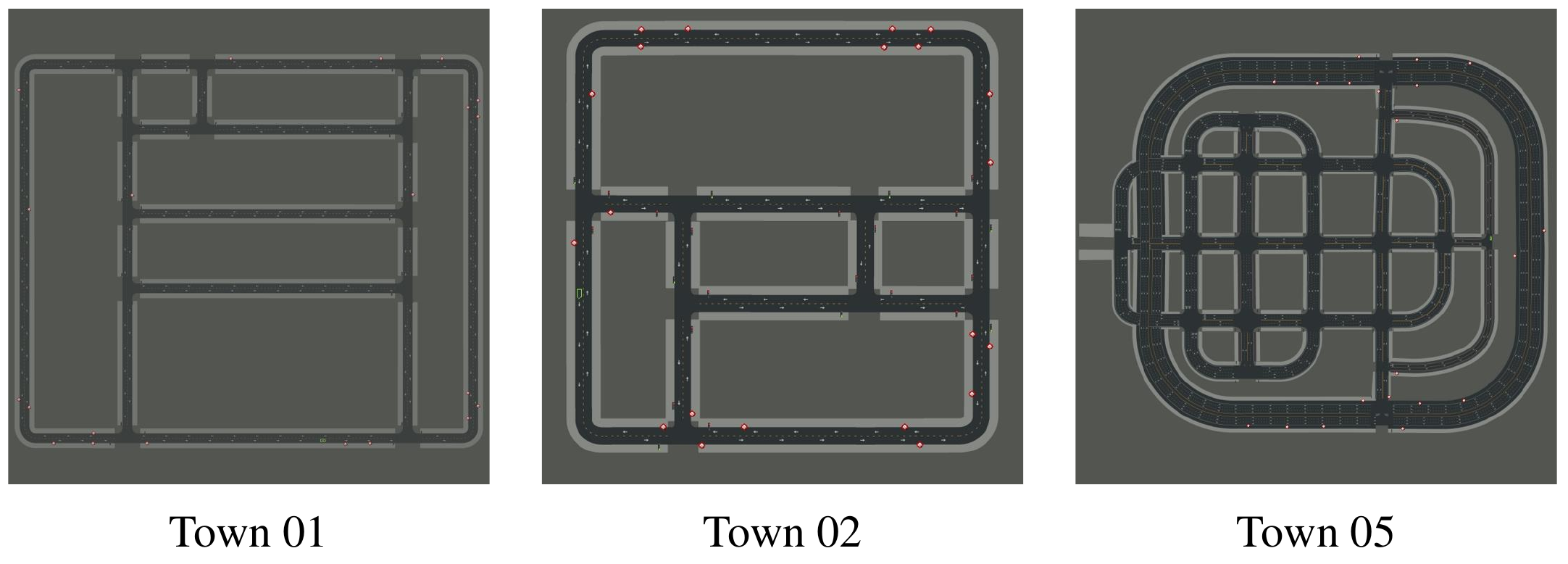}
    \caption{Maps of Town01, Town02, and Town05.}
    \label{fig:maps}
\end{figure}

\begin{figure}[tb]
    \centering
    \includegraphics[width=.65\textwidth]{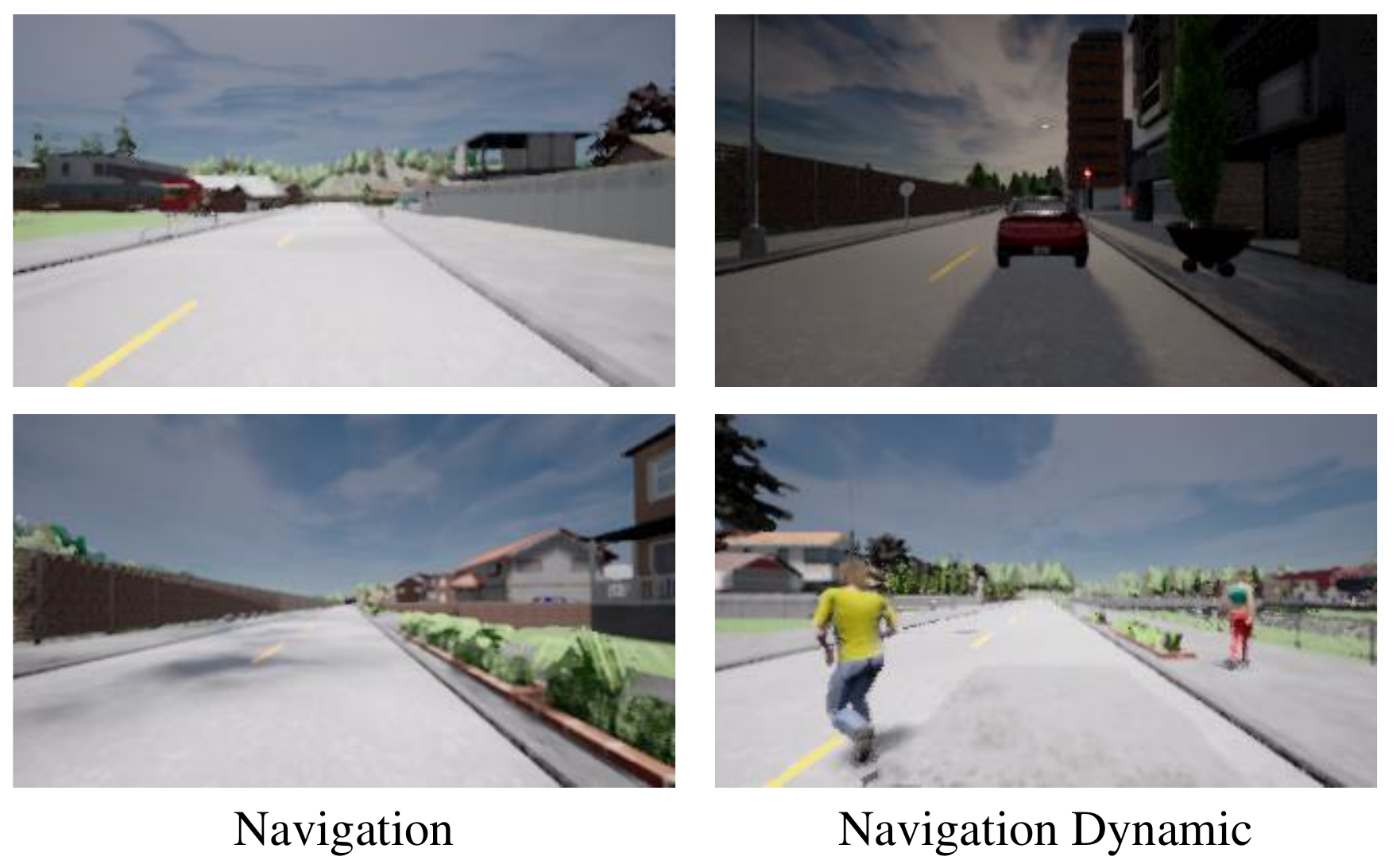}
    \caption{Examples of the front view image for Navigation and Navigation Dynamic tasks.}
    \label{fig:navigation-detail}
\end{figure}

For \textbf{Navigation} and \textbf{Navigation Dynamic}, training data is collected in Town01, and the closed-loop testing is conducted in Town02. The maps of Town01 and Town02 are shown in Fig.~\ref{fig:maps}. The agent needs to follow a series of sparse waypoints to navigate from the start point to the end point and avoid collisions. The difference between Navigation and Navigation Dynamic is that there are other dynamic vehicles and pedestrians in the town. Examples are provided in Fig.~\ref{fig:navigation-detail}.

The \textbf{Leaderboard-Town05-long} task is more close to real-world urban driving, with different challenging scenarios added to the route. The map of Town05 is shown in Fig.~\ref{fig:maps}. 


\section{Limitations} \label{sec:limitation}

\begin{figure}[htb]
    \centering
    \includegraphics[width=0.6\textwidth]{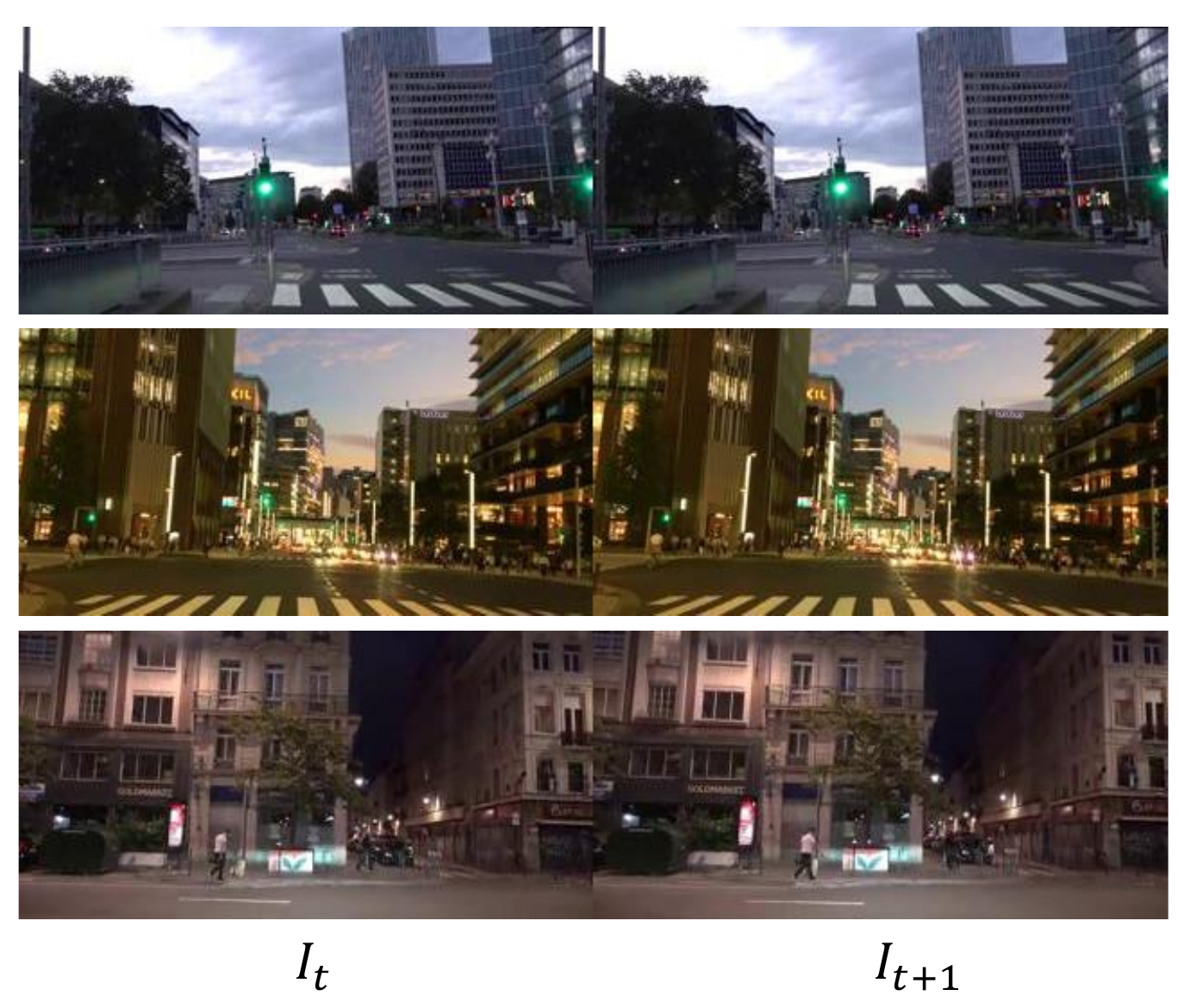}
    \caption{Failure cases where the driving decision/future motion can not be inferred from $I_t$. For the cases in Row 1 and Row 2, by comparing $I_t$ and $I_{t+1}$, we know that the ego vehicle stops. However, there is no clear clue in $I_t$ indicating it should stop. For the case in Row 3, the ego vehicle is turning left, while we could hardly tell the turning direction from $I_t$ alone.}
    \label{fig:failure_cases}
\end{figure}

In this part, we analyze some failure cases and limitations of our method. Since the visual encoder need to predict the future motion based on a single front-view image, there might be some factors that directly influence the driving decision not shown in the image (\textit{e.g.}, vehicles behind the ego vehicle, factors related to the driver, navigation information). Some of such cases are provided in Fig.~\ref{fig:failure_cases}. In these cases, the visual encoder does not get enough information to make the correct prediction. These samples during training may hamper the learning process. After training, one may use the difference between the  prediction from PoseNet and that from visual encoder to filter out these samples, and re-train the visual encoder.

\end{document}